\pdfoutput=1

\documentclass[11pt]{article}

\usepackage[preprint]{acl}

\usepackage{multirow}
\usepackage{times}
\usepackage{latexsym}
\usepackage{algorithm}
\usepackage{algorithmic}
\usepackage{amsmath} 
\usepackage{amssymb}
\usepackage[utf8]{inputenc} 
\usepackage{microtype}      
\usepackage{xcolor}         
\usepackage[T1]{fontenc}
\usepackage[T1]{fontenc}

\usepackage[utf8]{inputenc}

\usepackage{microtype}

\usepackage{inconsolata}

\usepackage{graphicx}

\title{Functional Abstraction of Knowledge Recall in Large Language Models}

\author{Zijian Wang \\
  School of Computer Science\\
  The University of Sydney \\
  \texttt{zwan0998@uni.sydney.edu.au} \\\And
  Chang Xu \\
  School of Computer Science\\
  The University of Sydney \\
  \texttt{c.xu@sydney.edu.au} \\}

\begin{document}
\maketitle
\begin{abstract}
Pre-trained transformer large language models (LLMs) demonstrate strong knowledge recall capabilities. 
This paper investigates the knowledge recall mechanism in LLMs by abstracting it into a functional structure.
We propose that during knowledge recall, the model's hidden activation space implicitly entails a function execution process where specific activation vectors align with functional components (Input argument, Function body, and Return values). 
Specifically, activation vectors of relation-related tokens define a mapping function from subjects to objects, with subject-related token activations serving as input arguments and object-related token activations as return values.
For experimental verification, we first design a patching-based knowledge-scoring algorithm to identify knowledge-aware activation vectors as independent functional components.
Then, we conduct counter-knowledge testing to examine the independent functional effects of each component on knowledge recall outcomes.
From this functional perspective, we improve the contextual knowledge editing approach augmented by activation patching. 
By rewriting incoherent activations in context, we enable improved short-term memory retention for new knowledge prompting.
\end{abstract}

\section{Introduction}

Pre-trained LLMs serve as knowledge bases \cite{petroni2019language, roberts2020much, }, efficiently storing and recalling diverse facts \cite{kassner-etal-2023-language}.
Given a subject-relation query, LLMs can predict the correct object \cite{jiang2020can}.
However, the internal mechanism of knowledge recall within LLMs still lacks transparency and interpretability, which limits LLMs from becoming reliable assistant tools \cite{hendrycks2022x, wang2023survey}.


There are many interpretability works trying to reveal how knowledge is organized inside the model.
The mainstream is module-based mechanical approaches, which typically attribute the storage and retrieval of knowledge to specific modules by identifying important model parameters (or neuron weights).
They treat sub-modules as basic building blocks and study how they perform different information processing steps, thus providing a low-level architectural perspective on knowledge recall \cite{wang2211interpretability, geva2023dissecting,merullo2023mechanism, geva2023dissecting, meng2022locating}.
However, the identification of salient modular components necessitates a substantial degree of manual effort, and it becomes difficult to interpret more complex phenomena using this framework \cite{lieberum2023does, hase2024does, cohen2024evaluating}.


\begin{figure}[t]
    \centering
    \includegraphics[scale=0.35]{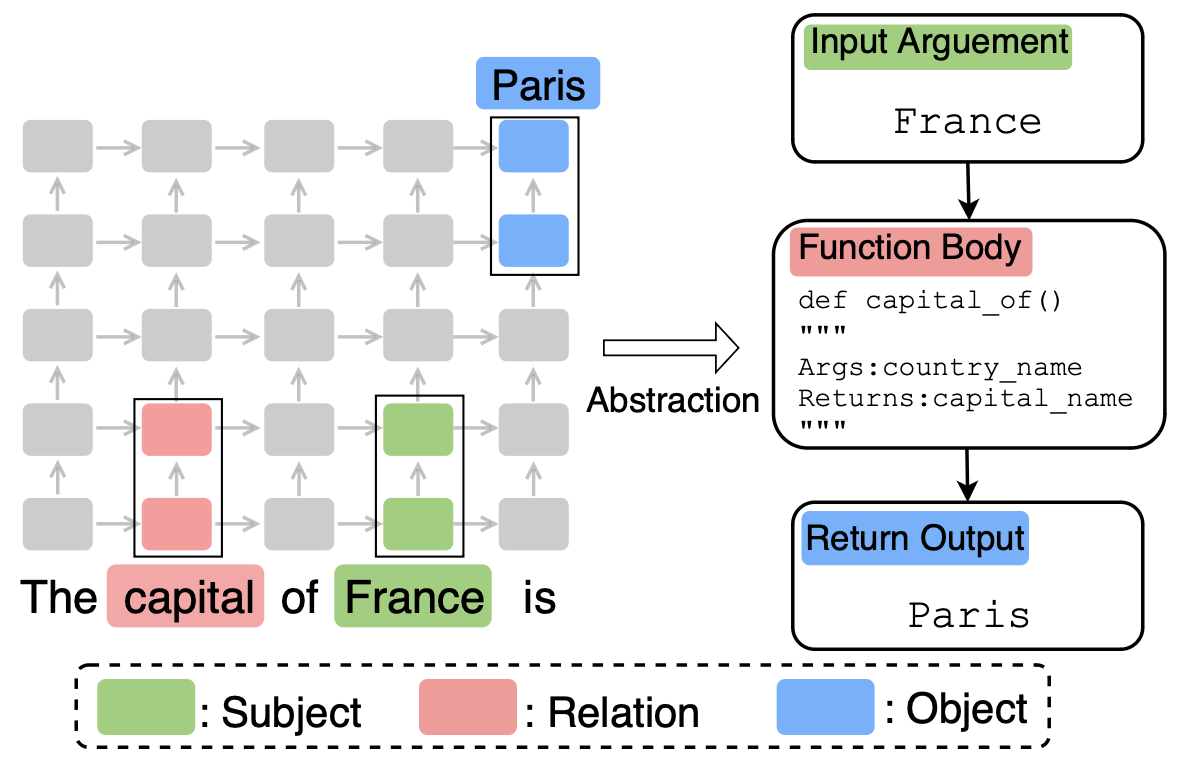}
    \caption{Illustration of our abstraction framework. In a knowledge recall process, \textit{i.e., (subject, relation)$\rightarrow$object}, we find that knowledge-related representations are locally distributed, and they are well-aligned with functional components.}
    \label{figure1}
\end{figure}

Another line of approaches is representation-based, which studies what kind of human-interpretable information the model has learned to represent within its activation space \cite{zou2023representation, von2024language}.
These methods consider the behavior of the model to be the product of the representation space, which is the result of the interaction of multiple representations \cite{li2024inference, burns2022discovering, wang2024locating}.
By identifying and locating interpretable representations, we can abstract the inner workings of the model into a well-defined algorithm, where the neural representations are aligned with the variables or components in the algorithm.

In this paper, we inherit the idea of representation analysis and design a three-step method to abstract the knowledge recall process of black-box LLMs into an interpretable functional structure.
The first step is to formulate a hypothesis, which aims to propose an interpretable algorithm that matches the model behavior.
We assume that in the forward propagation of knowledge recall, the definition and execution of a function are implicitly performed in the activation space \cite{hernandez2023linearity, todd2023function, hendel2023context} \footnote{In this paper, we use \textit{representation}, \textit{activation} and \textit{hidden state} interchangeably.}.
As shown in Figure \ref{figure1}, we use a function execution process to fit knowledge recall, which is that the subject is the input, the relationship is the mapping, and the object is the output. 
This is a reasonable assumption because the relationship is essentially a transformation operation between entities \cite{bordes2013translating}.

The second step is to locate representations that can be aligned with the algorithm variables.
To achieve that, we employ an activation patching technique \footnote{Also referred to as Causal Mediation Analysis, Causal Tracing, and  Interchange Interventions.} to probe interpretable representations.
Specifically, we aim to identify activation vectors that are particularly responsible for encoding information about the subject, relation, and object in the forward propagation, which is achieved through the knowledge-encoding scoring algorithm we design based on the insight of causal mediation analysis \cite{meng2022locating, vig2020investigating}.
This method can pinpoint knowledge-aware activation vectors by examining their knowledge-encoding scores. 
The results reveal a remarkable property that the distribution of relevant activation vectors is localized, which enables the isolation of these vectors and their treatment as distinct functional components (Input argument, function body, and return value).

Finally, the third stage is to experimentally verify that the localized neural representations exhibit properties that can be aligned as algorithmic variables.
We design rigorously controlled counterfactual testing experiments \cite{geiger2021causal, geiger2024finding} involving the construction of counter-knowledge data pairs and the execution of four distinct types of interchange intervention. 
By building counter-knowledge data pairs where we fix either the subject or relation and interchange the other one, we can isolate and independently assess their functional roles.
Our results support our hypothesis that these activation vectors function as distinct, separable components in the knowledge recall process.

Based on this functional perspective, we design a prompting-based knowledge editing method that does not require parameter tuning. 
When a new knowledge statement is presented, we resolve the conflict between the new knowledge and the LLM’s original knowledge by patching specific functional components. 
Subsequently, we use this patched knowledge statement as a context
prompt to inform the model that the knowledge has been edited. 

\section{Related Work}

\subsection{Mechanistic Interpretability Methods}
This type of approach, also called the modular method, aims to understand LLMs in terms of neurons, modules, or circuits from a mechanistic view \cite{dai2021knowledge, yao2024knowledge}.
They place model components as the analysis center, studying their algorithmic role that can execute specific functions for information processing, such as information storage and transmission.
\cite{meng2022locating} found that shallow-layer feed-forward module weights are responsible for subject knowledge storage.
\cite{geva2023dissecting, elhage2021mathematical} study shows the information transfer role of middle-layer attention module via attention knock-out.
\cite{wang2211interpretability,yao2024knowledge} discovered circuits composed of multiple distinct role components through reverse engineering, which can provide a bottom-up explanation of the model's behavior.

\subsection{Representational Interpretability Methods}
This type of approach mainly studies internal representations of LLMs, focusing on the decoding and mining interpretable information.
\cite{zou2023representation} viewed the model's behavior as the interaction of conceptual representations and found that some human-understandable concepts can be read from hidden states through feature dimensionality reduction methods. 
\cite{li2024inference} showed that the model exhibits different activation patterns for true and false statements and provided evidence that intervening in the hidden representations can mitigate hallucinations.
\cite{von2024language} utilized various linear probers to detect different conceptual representations effectively and provided linear guidance at activation space for text generation style control.

These above methods uncover the emerging high-level structure and characteristics that activation space can demonstrate while abstracting away intricate low-level mechanisms.

\subsection{Activation Patching for Interpretability}
The activation patching technique is a popular tool for identifying the specific activation responsible for particular functions. 
\cite{pearl2022direct}, which has been widely applied for LLMs interpretability \cite{zhang2023towards, hanna2024does, vig2020investigating}.

\cite{meng2022locating, geva2023dissecting} used this method for knowledge-storing modules localization.
\cite{wang2211interpretability, hanna2024does, lieberum2023does, conmy2023towards} identified the sub-network within a model’s computation graph that implements a specified behavior. 
\cite{zhang2023towards} systematically investigated the prompt corruption method and metrics for patching evaluation.
\cite{ghandeharioun2024patchscope} decoded information from hidden representations by generating human-interpretable texts from post-patching computation.

In this work, we apply this technique to locate the knowledge-aware representations and validate the alignment of these representations with a pre-hypothesized functional structure.

\begin{figure}[t]
    \centering
    \includegraphics[scale=0.50]{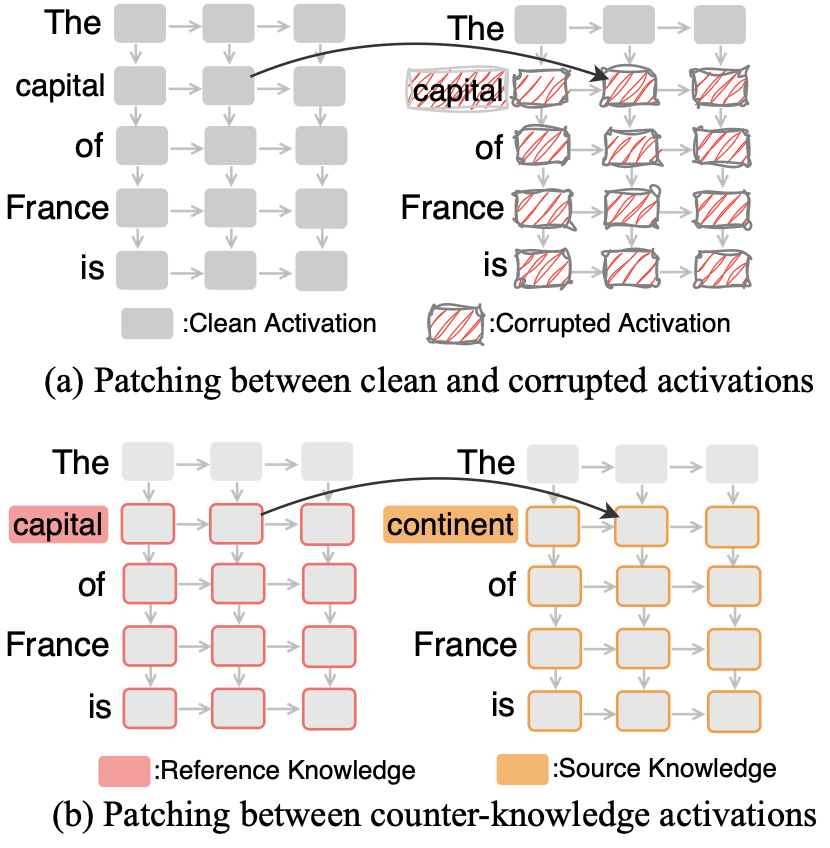}
    \caption{Illustration of two activation patching operations. 
    In (a), we get the corrupted activations by adding noise to embeddings and then patch the corrupted ones with clean activations. 
    In (b), we patch the source knowledge with the activations from the reference knowledge.}
    \label{figure2}
\end{figure}

\section{Method}
Given the reasonable assumption that knowledge recall is a function execution process, in this section, we introduce how to identify the knowledge-related representations through the proposed knowledge-scoring algorithm, as well as how to validate their alignment with functional components through the counter-knowledge test.

\subsection{Notions}
In this paper, we consider a knowledge recall scenario where an LLM $\mathcal{T}$ takes a subject ($s$)--relation ($r$) query as input and predicts the relevant object as the continuation. 
Using the template \textit{"The <relation> of <subject> is"}, the query is tokenized into an embedding sequence $I(s,r) = [\tau_1, \tau_2,...,\tau_n]$, where $n$ is the token length. 
We denote the token positions for the subject and relation as $S$ and $R$, respectively. 
$I(s,r)$ is processed by a multi-layer transformer, generating activation vectors at each layer. 
The model outputs a probability $p(o| I(s, r))$ through a classification head $\phi$. We use $h_{l,j}$ to represent the activation vector at layer $l \in [0 ,..., L-1]$ and token position $j\in [0,...,n-1]$.


\begin{align*}
     P(o|I(s,r))= & \mathrm{Softmax}(\phi(h_{L-1, n-1})) \\
     h_{l, j} =& h_{l-1, j} + \mathrm{Attn}(h_{l-1, \leq j }) + \\ 
     &\mathrm{MLP}(\mathrm{Attn}(h_{l-1, \leq j}), h_{l-1, j}) \\
\end{align*}

\subsection{Knowledge Scoring via Activation Patching}
\paragraph{Overview}

The basic idea is inspired by causal mediation analysis \cite{meng2022locating, meng2022mass}, which has been used to identify knowledge-storing parameters by only analyzing subject concepts.
In this work, we use this technique to identify the activation vectors that encode specific knowledge by treating the activation vectors as indirect causal variables and calculating their mediating effects as knowledge-encoding scores.
A high score suggests that the activation vector effectively represents rich and relevant knowledge.

\begin{figure*}[t]
    \centering
    \includegraphics[scale=0.62]{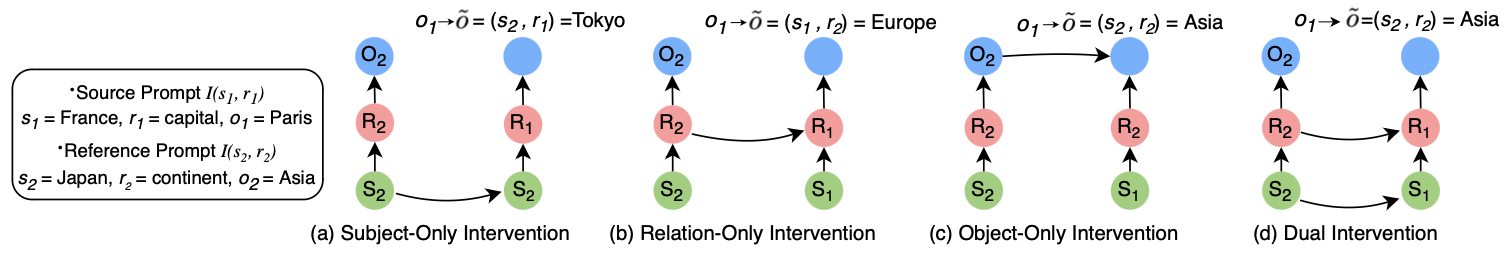}
    \caption{Illustration of the counter-knowledge testing experiments.
    If the prediction of the object after the interchange is affected only by the exchanged knowledge, it can be proved that each knowledge vector acts as a functional component independently.}
    \vspace{-0mm}
    \label{figure3}
\end{figure*}

The score calculation is a two-step process, including textual knowledge ablation and corruption restoration.


\paragraph{Textual Knowledge Ablation}

We achieve subject ablation, relation ablation, and object ablation, respectively, through noisy embeddings.

To obtain the subject ablation embeddings $I(s^*, r)$, we first create the noised subject token embeddings $\tau^*_S$, \textit{i.e.}, $\tau^*_S = \tau_S + \epsilon$, where $\epsilon$ is Gaussian noise. 
Then, the $\tau^*_S$ continues to propagate forward, causing the following activation vectors to be corrupted.  
We use $h^{s*}_{l,j}$ to represent the activation vectors that are corrupted by the presence of $\tau^*_S$.

Similarly, the relation ablation embeddings $I(s, r^*)$ and the following corrupted $h^{r*}_{l,j}$ can also obtained by noising the relation embeddings, \textit{i.e.}, $\tau^*_R = \tau_R + \epsilon$. 

To ablate the object knowledge in the text query, we choose to add noise to both the subject and object embeddings simultaneously.
In this way, we obtain the object ablation embeddings $I(s^*, r^*)$ and the relevant $h^{o*}_{l,j}$.

\paragraph{Corruption Restoration}
With these there types of knowledge ablation embeddings, we can calculate the Subject-Encoding score (SES), Relation-Encoding Score (RES), and Object-Encoding Score (OES) through corruption restoration, achieved by patching corrupted activations with clean ones.

Specifically, the three scores of each activation $h_{l,j}$ are calculated in three model runs.

$\bullet$ Run the model $\mathcal{T}$ on standard clean input embeddings $I(s, r)$ and all normal clean activation vectors $h_{l,j}$ are cached.

$\bullet$ Run the model $\mathcal{T}$ on the three ablated embeddings $I(s^*, r), I(s, r^*), I(s^*, r^*)$ respectively and the corrupted probabilities $P(o| I(s^*, r))$, $P(o| I(s, r^*))$ and $P(o| I(s^*, r^*))$ are obtained.

$\bullet$ Run the model $\mathcal{T}$ on the ablated embeddings again and patch each corrupted activation vector with clean ones one by one.
Formally, we use $\mathrm{do}(h^{s*}_{l,j}=h_{l,j}) = h^{s*}_{l,j} \leftarrow h_{l,j}$ to denote the patching operation, where the corrupted $h^{s*}_{l,j}$ is replaced by a clean $h_{l,j}$. 
Then, the restored probabilities $P(o | I(s^*, r), \mathrm{do}(h^{s*}_{l,j} = h_{l,j}))$, $P(o | I(s, r^*), \mathrm{do}(h^{r*}_{l,j} = h_{l,j}))$ and $P(o | I(s^*, r^*), \mathrm{do}(h^{o*}_{l,j} = h_{l,j}))$ are obtained.

\paragraph{Score Calculation}
The three knowledge-encoding scores of an activation vector $h_{l,j}$ are defined as the difference between the restored and the corrupted probability.
\begin{align*}
SES(h_{l,j}) &= P(o | I(s^*, r), h^{s^*}_{l,j} \leftarrow h_{l,j}) \\
             &\quad - P(o| I(s^*, r)) \\
RES(h_{l,j}) &= P(o | I(s, r^*), h^{r^*}_{l,j} \leftarrow h_{l,j}) \\
             &\quad - P(o| I(s, r^*)) \\
OES(h_{l,j}) &= P(o | I(s^*, r^*), h^{o^*}_{l,j} \leftarrow h_{l,j}) \\
             &\quad - P(o| I(s^*, r^*))
\end{align*}

Through comparative analysis, we can locate activation vectors with high scores and treat them as subject, relation, and object representation vectors.

\subsection{Counter-Knowledge Testing}
Given the identified subject, relation, and object representations, we treat them as functional components and assess the alignment through counter-knowledge data preparation and activation interchange.


\paragraph{Counter-knowledge Data} 
By constructing such source-reference pairs, we can interchange the subject or relation representations to disentangle their individual functional role in object prediction.
Each counter-knowledge data pair consists of a source query $I(s_1,r_1)$ with object $o_1$ and a reference query $I(s_2,r_2)$ with object $o_2$, where either the subjects or relations differ ($s_1 \neq s_2, r_1 = r_2$ or $s_1 = s_2, r_1 \neq r_2$ ), or both ($s_1 \neq s_2$ and $r_1 \neq r_2$ ).

\paragraph{Activation Vectors Interchange}

Activation vector interchange replaces a specific activation vector from a model's source run with one from a reference run to intervene in knowledge recall.
By observing changes in object predictions $\tilde{o}$ after the swap, we can determine if each knowledge vector functions independently.
If the predictions change only with the reference vector, it indicates that the vectors operate as independent functional components.

Given a source-reference pair, $I(s_1, r_1)$  and $I(s_2, r_2)$, we perform activation interchange between the two model runs. Let $\beta_{S1}$, $\beta_{R1}$, and $\beta_{O1}$ represent the source subject, relation, and object vectors, and $\beta_{S2}$, $\beta_{R2}$, and $\beta_{O2}$ represent the reference ones. 
Four experimental settings are designed based on different interchange sources and targets, as shown in Figure \ref{figure3}.



1. Subject-only Intervention. 
Only replace the source subject vector with the reference one to get object probability $P_s(\tilde{o} | I(s_1, r_1), \mathrm{do}(\beta_{S1} = \beta_{S2}))$, where $\tilde{o} = (s_2, r_1)$.

2. Relation-only Intervention. 
Only replace the source relation vector with the reference one to get object probability $P_r(\tilde{o} | I(s_1, r_1), \mathrm{do}(\beta_{R1} = \beta_{R2}))$, where $\tilde{o} = (s_1, r_2)$.

3. Object-only Intervention. 
Only replace the source object vector with the reference one to get object probability $P_o(\tilde{o} | I(s_1, r_1), \mathrm{do}(\beta_{O1} = \beta_{O2}))$, where $\tilde{o} = (s_2, r_2)$.

4. Dual Intervention. Replace both the source subject and relation vectors with reference ones to get object probability $P_d(\tilde{o} | I(s_1, r_1), \mathrm{do}(\beta_{S1} = \beta_{S2}),\mathrm{do}(\beta_{R1} = \beta_{R2}))$, where $\tilde{o} = (s_2, r_2)$. 
This reflects an architectural composition of activation vectors for knowledge recall.

\subsubsection{Evaluation}
We use two metrics to evaluate whether an individual knowledge vector performs its functional role independently:

1. Interchange Effect Comparison. 
We respectively calculate the interchange effects on the text knowledge and on the vector knowledge, \textit{i.e.},
$P_s(\tilde{o} | I(s_1, r_1), \mathrm{do}(\beta_{S1} = \beta_{S2}))$ and $P_s(\tilde{o} | I(s_2, r_1))$, where $\tilde{o} = (s_2, r_1)$.

2. Interchange Accuracy. 
We calculate the matching success rate between the prediction results after text knowledge intervention and the prediction results after representation knowledge intervention, \textit{i.e.}, $\mathbb{E}[\operatorname{argmax}_{\tilde{o}} P_s(\tilde{o} | I(s_1, r_1), \mathrm{do}(\beta_{S1} = \beta_{S2})) \stackrel{?}{=} \operatorname{argmax}_{o} P(o| I(s_2, r_1))]$.

\section{Experiments}
\paragraph{Model}
We choose Llama2-7b \cite{touvron2023llama} and Mistral-7b \cite{jiang2023mistral} as our models, each one with 32 layers. 
We use huggingface implementations \cite{wolf2020transformers} of each model and run them on an A100 40GB GPU.

\paragraph{Data}
We utilize the relational knowledge database from \cite{hernandez2023linearity}, which consists of 26 types
 of factual knowledge.
To ensure that the knowledge has been stored in the model, we filter the dataset to examples where the model correctly predicts the object $o$ given the query.
More details are in the appendix.

\paragraph{Query templates}
Given the existence of unidirectional self-attention in LLMs, which only grants subsequent tokens access to previous tokens, the order of subject and relation occurrence within the query may result in semantic entanglement of relevant activation vectors.
Therefore, to verify that the subject-relation order does not affect their functional roles, we designed two types of query templates with different subject-relation orders, that are \textit{"The <relation> of <subject> is"} and \textit{"Given <subject>, it's <relation> is"}.

\begin{figure}[t]
    \centering
    \includegraphics[scale=0.315]{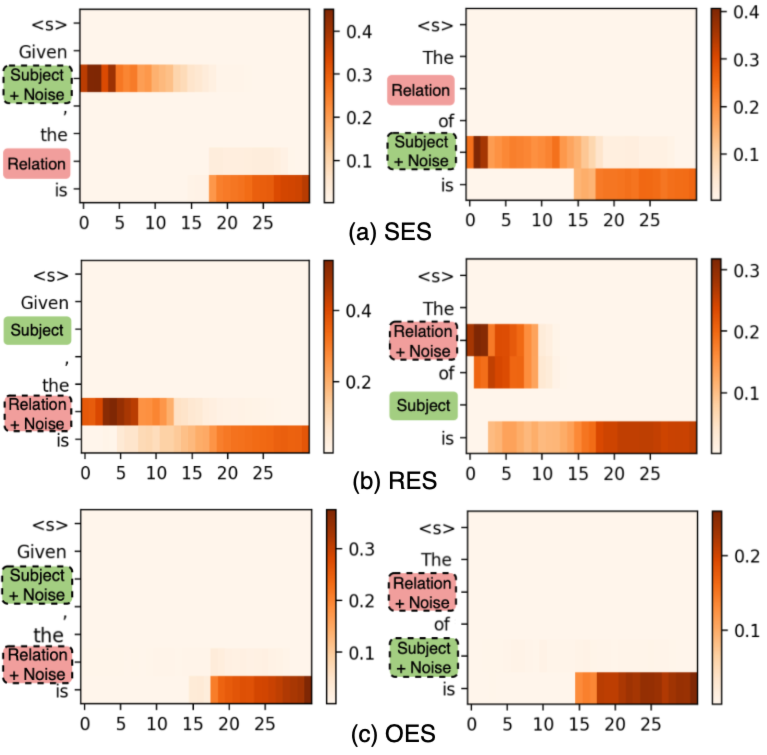}
    \caption{Scores heat map visualization. The left column uses the template \textit{"Given <subject>, the <relation> is"}, and the right column uses  \textit{"The <relation> of <subject> is"}.}
    \label{figure4}
\end{figure}

\subsection{Scoring Reveals Localized Features}
Figure \ref{figure4} shows the distribution map of three types of knowledge scores obtained from two prompt templates with different subject-relation orders.
We can clearly observe the locality of the distribution of the corresponding activation vectors.
We conduct more detailed locality analyses along the position dimension and the layer dimension, respectively.

\subsubsection{Locality Along Token Positions}
\begin{figure}[h]
    \centering
    \includegraphics[scale=0.42]{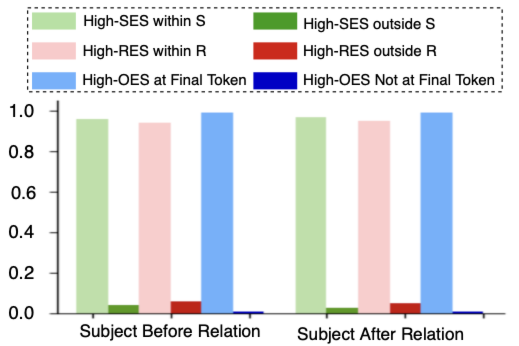}
    \caption{Locality along token positions. This bar graph shows the proportion of high-score (>0.05) activation vectors in different token ranges.}
    \label{figure5}
\end{figure}

In Figure \ref{figure5}, we demonstrate the locality of the high-scoring activation vectors along the token position dimension.
We find that the vectors with high OES scores are all located at the last token. 
This result is reasonable, as the classification head will use the activation vector of the last token to predict the next token.
Early decoding studies have also discussed similar findings that project activation vectors into vocab space in advance to examine internal information \cite{ghandeharioun2024patchscope, hernandez2023linearity}.

Interestingly, the vectors with high SES and high RES are almost located within the range of the subject and the relation tokens, respectively, and the order of the subject-relation does not seem to matter, although the self-attention is unidirectional.
This observation reveals that the language model does not encode knowledge in a distributed manner but rather confines the knowledge representations to those relevant tokens.

\subsubsection{Locality Along Layer Positions}
\begin{figure}[h]
    \centering
    \includegraphics[scale=0.29]{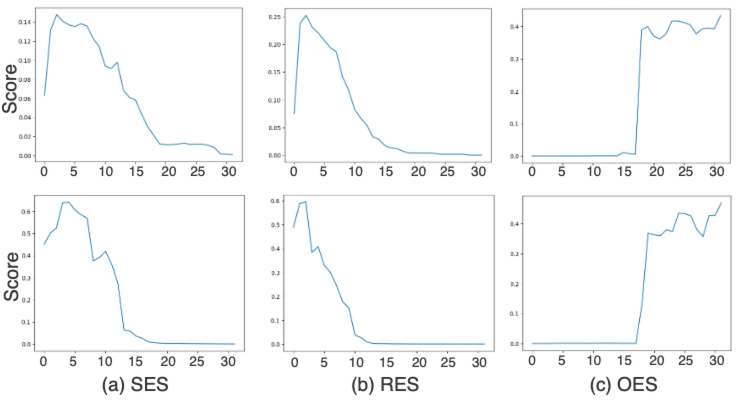}
    \caption{Locality along layer positions. The plots of average SES, RES, and OES at the identified token positions, respectively, where the upper template is \textit{"Given <subject>, the <relation> is"} and the lower template is \textit{"The <relation> of <subject> is"}.}
    \label{figure6}
\end{figure}

In Figure \ref{figure6}, 
we plot line graphs showing how the scores change across the layers, focusing on their relevant tokens.
We find that the vectors with high OES scores are located in the middle to late layers, while the vectors with high SES and RES scores are located in the early to middle layers.
This finding is somewhat consistent with the explanation based on information aggregation, that is, the information of the subject is transferred in the middle layers \cite{geva2023dissecting}.

\subsubsection{Formulating Functional Components}
Based on the above findings, we define activation vectors that can be treated as components of a knowledge recall function.

$\bullet$ Subject Vectors: Activation vectors located in early layers (0-14) within the subject tokens range, serving as input arguments.

$\bullet$ Relation Vectors: Activation vectors located in early layers (0-10) within the relation tokens range, serving as the function body.

$\bullet$ Object Vectors: Activation vectors located in late layers (15-31) in the final token, serving as return values.

\subsection{Counter-Knowledge Testing}
In this experiment, we present the results of the counter-knowledge testing to demonstrate that the identified knowledge vectors operate in a functional manner to recall the knowledge.

\begin{figure}[h]
    \centering
    \includegraphics[scale=0.24]{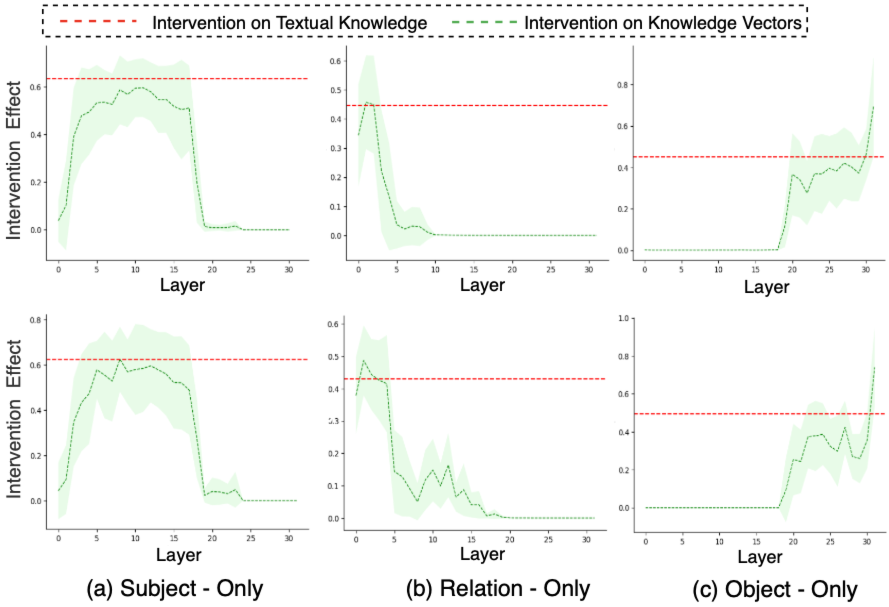}
    \caption{The plots of intervention effects of textual and representational knowledge, respectively. The upper prompt template is \textit{"Given <subject>, the <relation> is"} and the lower prompt template is \textit{"The <relation> of <subject> is"}.}
    \label{figure7}
\end{figure}

\paragraph{Layer-by-layer Interchange}
In this setting, we interchange the corresponding knowledge vectors layer by layer and record the intervention effects after each operation.
We also calculate the intervention effect of textual knowledge interchange across all relation types.
The results are shown in Figure \ref{figure7}. 
We plot the intervention effect in subject-only, relation-only, and object-only settings, where the green line represents the intervention effect on knowledge vectors, and the red line represents the intervention effect on textual knowledge.
We can observe that the effect of subject-only and relation-only intervention is obvious in the early to middle layers, and the effect of object-only intervention is obvious in the middle to late layers. 
This observation is consistent with the previous knowledge encoding scoring results, suggesting that our designed scoring algorithm accurately identifies knowledge-aware representations.

We can also notice that the intervention on hidden states achieves results comparable to those of the intervention on textual knowledge.
This indicates that the intervened hidden states faithfully encode knowledge and that these hidden states and causal variables are well aligned.

\paragraph{Mean Interchange}
In this setup, we replace the knowledge vectors in the source model run with the corresponding mean reference knowledge vector \footnote{We extract the mean of the vectors at the corresponding token positions and layer from the reference model run.} and then calculate the prediction accuracy of the target after the interchange.
The exchange accuracy of subject and relation vectors also decreased slightly.

As shown in Table \ref{table1}, the results show that the accuracy of object-only vector interchange is almost 100\%, suggesting that the object vector exhibits a high degree of fidelity as the function return value.
The accuracy of the subject-only and the relation-only vector interchange also decreased slightly.
This shows that the mean knowledge vector we extracted fully and independently encodes the corresponding knowledge representation.
In the dual interchange scenario, the accuracy also drops slightly, indicating that each knowledge vector can be used as an individual function component to complete knowledge recall.

\begin{table}[h]

\label{table1}
\resizebox{0.5\textwidth}{!}{%
\begin{tabular}{cllll}
\hline
\multicolumn{1}{l}{} & \multicolumn{1}{c}{\begin{tabular}[c]{@{}c@{}}Subject-\\ Only\end{tabular}} & \multicolumn{1}{c}{\begin{tabular}[c]{@{}c@{}}Relation-\\ Only\end{tabular}} & \multicolumn{1}{c}{\begin{tabular}[c]{@{}c@{}}Object-\\ Only\end{tabular}} & Dual   \\ \hline
\multicolumn{5}{l}{\textit{The <relation> of <subject> is}}        \\ \hline
Country-Capital    & 97.1\%   & 93.6\%    & 100.0\%    & 91.1\% \\
Food-Country       & 96.4\%   & 93.5\%    & 100.0\%    & 92.5\% \\
Company-Headquarter& 93.6\%   & 93.1\%    & 99.4\%    & 91.3\% \\
Product-Company    & 92.4\%   & 92.9\%    & 99.1\%    & 89.2\% \\
Landmark-Country   & 93.8\%   & 92.3\%    & 100.0\%    & 91.7\% \\
Fruit-Color        & 94.1\%   & 93.7\%    & 100.0\%    & 92.5\% \\ \hline
\multicolumn{5}{l}{\textit{Given <subject>, its <relation> is}}                                                                                                                                                                                                                      \\ \hline
Country-Capital      & 96.5\%   & 94.8\%    & 100.0\%   & 92.7\% \\
Food-Country         & 97.1\%   & 93.2\%    & 100.0\%    & 91.8\% \\
Company-Headquarter  & 96.6\%   & 93.9\%    & 99.5\%    & 88.5\% \\
Product-Company      & 97.3\%   & 94.8\%    & 99.3\%   & 89.2\% \\
Landmark-Country     & 95.9\%   & 93.4\%    & 100.0\%    & 90.8\% \\
Fruit-Color          & 95.3\%   & 94.9\%    & 100.0\%    & 91.4\% \\ \hline
\end{tabular}
}
\caption{The Mean Interchange Accuracy across different relations, using two prompts with different subject-relation order, respectively.}
\label{table1}
\end{table}

\begin{figure}[ht]
    \centering
    \includegraphics[scale=0.6]{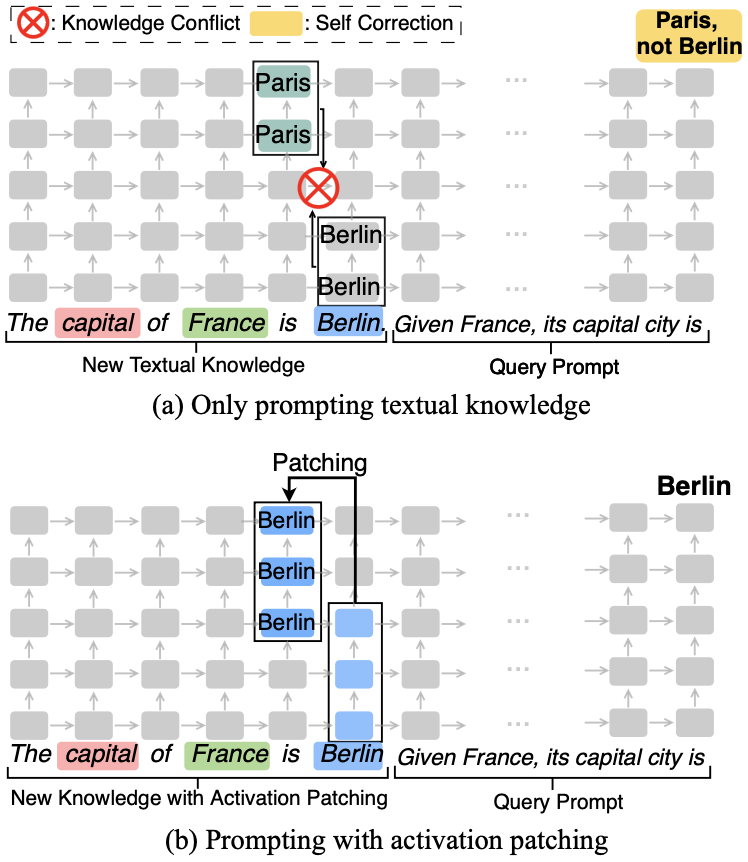}
    \caption{Figure (a) shows the knowledge conflict in contextual knowledge editing. Figure (b) shows our proposed activation patching-based knowledge editing method. }
    \label{figure8}
\end{figure}

\section{Functionality Informs Knowledge Editing}
Given the credibility of the implicit functional structure in knowledge recall, it is natural to think about how to apply it to knowledge editing, which aims to insert a new knowledge triplet $(s, r, o^*)$ in place of the original triplet $(s, r, o^c)$. 
In this section, we improve the contextual knowledge editing method based on functional activation patching.

Contextual knowledge editing provides short-term memory of new knowledge by appending context prompting before queries \cite{madaan2022memory, zheng2023can, yin2024history}.
However, from a functional perspective, the new knowledge prompting function implicitly returns an object that conflicts with the new object, causing the model to produce self-correcting behavior.

As shown in Figure \ref{figure8} (a), when a new knowledge prompt \textit{"The capital of France is Berlin"} is fed to the model, the original object \textit{"Paris"} conflicts with the new object \textit{"Berlin"} in the activation space.
This discrepancy triggers the model's awareness of the inconsistency between the new knowledge and its existing knowledge. 
Consequently, the model attempts to self-correct its predicted answer to align with its original knowledge.

\subsection{Method}

To dissolve the knowledge conflict, we patch the original object representation with the new object vector.
Specifically, as illustrated in Figure \ref{figure8} (b), during the forward pass of the new knowledge prompt, we replace the late layer activation vectors of the previous token of objects, \textit{e.g.}, the \textit{"is"} token, which has been proven to encode the original object information, with the new object mean activation vector. 
In this way, the context prompting completely implements the recall function of new knowledge and better provides short-term memory without knowledge conflict.
We extract the mean activation vector at the early layers (0-14) from the object tokens of new knowledge prompting to obtain the new object vector. 

\subsection{Data}
We construct 3000 new knowledge triplets to evaluate our method by replacing the objects of the original triplets.
Then, we fill each new knowledge triplet into the template to get knowledge statements for context prompting.
We design two types of templates:

(1) Declarative Format: Directly uses affirmations to state new facts, which is concise and suitable for knowledge expression. 
We use the previous two templates: (1) \textit{"The <relation> of <subject> is <object>."} and (2) \textit{"Given <subject>, it's <relation> is <object>."}.

(2) Q\&A Format: Simulates the process of Q\&A, first posing a question and then providing the corresponding answer, which is more interactive \cite{zheng2023can}.
We use the prompt \textit{"Q: Tell me the <relation> of <subject>. A: <object>."}.


\subsection{Results}
We use the Efficacy Score and Efficacy Magnitude to measure the performance.
The Efficacy Score (ES) quantifies the accuracy that the edited output $o^*$ is more accurate than the original output $o^c$, defined as: $\mathbb{E}\left[\mathbb{I}\left[\mathcal{P}\left(o^*\right)>\mathcal{P}\left(o^c\right)\right]\right]$.
The Efficacy Magnitude (EM) measures the average increase in probability due to editing, defined as:
$\mathbb{E}\left[\mathcal{P}\left(o^*\right)-\mathcal{P}\left(o^c\right)\right]$.



In Table \ref{table2}, we can observe that activation patching
can significantly improve ES and EM regardless of the prompt template.
\begin{table}[h]
\setlength{\tabcolsep}{2mm}
\resizebox{0.5\textwidth}{!}{
\begin{tabular}{lcc}
\hline
\multicolumn{1}{c}{Input types}   & ES & EM \\ \hline
Query-Only                        & 0.000            & -0.230   \\
Declarative1 + Query              & 0.171            & 0.058   \\
Declarative1 (Patching) + Query   & \textbf{0.664}   & \textbf{0.286}   \\
Declarative2 + Query              & 0.143            & 0.132   \\
Declarative2 (Patching) + Query   & \textbf{0.657}   & \textbf{0.314}  \\
Q\&A + Query                      & 0.192            & 0.231   \\
Q\&A (Patching)+ Query            & \textbf{0.685}   & \textbf{0.381}   \\ \hline
\end{tabular}
}
\caption{The Efficacy Score and Magnitude of knowledge editing using different input baselines.}
\label{table2}
\end{table}

\section{Conclusion}
In this paper, we design a two-stage approach to reveal the underlying functional structure of the knowledge recall process in LLMs. 
In the first stage, we propose a knowledge-encoding scoring algorithm to identify isolated knowledge-aware activation vectors and treat them as independent function components.
In the second stage, we construct counter-knowledge data pairs and verify the alignment of knowledge vectors with functional components by activation interchange interventions.
Experimental results reveal that knowledge vectors are highly faithful to their functional roles.
We also design a contextual knowledge editing algorithm based on the function structure and used activation patches to achieve more efficient knowledge prompts.

\section{Limitation}
Although this paper conceptualizes the knowledge recall of LLMs as a functional structure, several limitations persist. Foremost, the proposed knowledge scoring algorithm relies on a brute-force search method, resulting in high time complexity. There is significant potential in developing more efficient knowledge exploration algorithms.

Furthermore, our framework awaits validation on text-generation and general question-answering tasks that specifically involve knowledge recall. The accuracy of the proposed knowledge editing method also requires improvement, suggesting that representation rewriting does not fully encapsulate the original knowledge. Moving forward, a deeper exploration of the knowledge storage mechanisms within LLMs is necessary.

\section{Acknowledgement}
This work was fully funded by BrewAI through a PhD program.
We would like to express our sincere gratitude to BrewAI and its CEO, Gavin Whyte, for their invaluable support of this research project. 
BrewAI is a leading company in the field of Generative Artificial Intelligence, dedicated to providing automated AI services to companies. 
Their expertise and commitment to research innovation provided a critical foundation for this research project.

\bibliography{custom}

\appendix

\section{Appendix}
\label{sec:appendix}

\subsection{Experimental Details}

\subsubsection{Data Preprocessing}
In this paper, we filter the relational knowledge data \cite{hernandez2023linearity} and select the data whose relevant knowledge is already stored in the model.
Specifically, each piece of knowledge data is represented by a triple (subject, relation, object). 
We fill the subject and relation into the predefined query template to predict the corresponding object.
We filter data with predefined templates of different subject-relation orders, which are \textit{"The <relation> of <subject> is" } and \textit{"Given <subject>, its <relation> is"}.
If the object can be successfully predicted by these two templates both, it can be considered that this knowledge has been stored in the model.
Table 1 shows the amount of data before and after filtering across various relation types.
\begin{table}[h]
\caption{The initial quantity and filtered quantity of each relation type. The filtered objects can be predicted as the next token by the prompt \textit{"Given $<$subject$>$, the $<$relation$>$ of this one is"}}
\label{data_details}
\resizebox{0.5\textwidth}{!}{
\begin{tabular}{llll}
\hline
\multirow{2}{*}{Subject-Object} & \multirow{2}{*}{Initial quantity} & \multicolumn{2}{c}{Filtered quantity} \\
                                &            & Mistral-7b       & Llama-7b      \\ \hline
Country-Capital                 & 253        & 155              & 168                \\
Country-language                & 24         & 14               & 18               \\
Landmark-Country                & 836     & 531              & 547                \\
Country-Currency                & 30        & 25               & 26                 \\
Product-Company                 & 522       & 278              & 289                \\
Company-Headquarters            & 674     & 348              & 351                \\
Food-Country                    & 30     & 22               & 24                 \\
Food-Color                      & 30      & 15               & 19                 \\
City-Country                    & 27       & 27               & 24                 \\\hline
\end{tabular}
}
\end{table}

Our subsequent knowledge scoring and counter-knowledge test experiments are all conducted on this filtered dataset.

\subsection{Knowledge Scoring}
In this section, we introduce the technical details of the knowledge-scoring experiment.
\subsubsection{Textual Knowledge Ablation}
We add Gaussian noise $\mathcal{N}(0, v)$ to the embeddings of certain tokens, where $v$ is 5 times the standard deviation of the token embeddings from all data.
If the subject and relation are tokenized into multiple tokens, we add noise to these tokens at the same time.

\subsubsection{Score Calculation}
When calculating the object prediction probability, since the object is predicted in the form of question completion, we only consider the probability of the first token predicted.

\subsection{Counter-Knowledge Testing}

\subsubsection{Data Pair Construction}
When constructing source-reference prompt pairs from a knowledge triple, we choose different strategies depending on the components being interchanged.

When performing the subject-only interchange, we fill the reference template with a subject from the triple and another unrelated relation, and fill the source template a the relation from the triple and another unrelated subject.

When performing the relation-only interchange, we fill the reference template with a relation from the triple and another unrelated subject, and fill the source template with a subject from the triple and another unrelated relation.

When performing the object-only and dual interchange, we fill the reference template with the a relation and a subject from the triple, and fill the source template with unrelated relations and subjects.

\subsubsection{Activation Vectors Interchange}
The interchange operation is divided into two steps: extracting and replacing knowledge vectors.

When extracting the subject and relation vectors, we extract the hidden activation vector at the last token position of the subject and relation from the reference process.
When extracting the object vector, we extract the hidden activation vector at the last token position of the entire query prompt.

When performing subject replacement, we replace the hidden activations of all subject tokens in the source prompt with the corresponding reference subject vector.
When performing relation replacement, we replace the hidden activations of all relation tokens in the source prompt with the corresponding reference relation vector.
When performing object replacement, we replace the hidden activation of the last token in the source prompt with the corresponding reference object vector.

When calculating the interchange effect and accuracy, we only consider the probability that the first token predicted belongs to the target object.

\subsection{Knowledge editing Data}

For each factual knowledge triple, we replace the object to construct a new triple, and fill the new triple into the template to get new knowledge. 
The template is the first declarative format \textit{"The <relation> of <subject> is <object>"}.
For the following query template, we randomly select from the declarative format 2 and the QA template.

\end{document}